\documentclass[10pt]{article}

\usepackage{arxiv}

\usepackage[utf8]{inputenc}
\usepackage[T1]{fontenc}
\usepackage[hyphens]{url}
\usepackage{graphicx}
\usepackage{booktabs}
\usepackage{multirow}
\usepackage{amsmath}
\usepackage{amssymb}
\usepackage{amsthm}

\usepackage{float}
\usepackage{placeins}
\usepackage[round]{natbib}
\usepackage{caption}
\usepackage[colorlinks=true, allcolors=blue, breaklinks=true]{hyperref}
\graphicspath{{figs/}}

\title{Beyond Cultural Persona Bias: Response Geometry as a Measure for Language Models and People}
\shorttitle{Beyond Cultural Persona Bias}
\headeright{Preprint}

\author{%
  Yuan Yuan\\
Independent Researcher, San Mateo, USA\\
Corresponding author: yzy0014@auburn.edu, https://yzy0014.github.io/\\
}
\date{\today}

\begin{document}
\maketitle

\begin{abstract}
Researchers are increasingly prompting language models with cultural personas so that they answer questionnaires as diverse human groups would. However, studies often report that these persona responses do not predict the models' behavior. We argue that whether the persona is Western or non-Western, the model's answers carry a cultural preference, and that this preference masks the variables that could predict its behavior.

We examine the prompted preference at two levels: aggregate Big Five scores and response geometry, the correlation matrix between dimensions within each respondent, represented as a symmetric positive-definite (SPD) matrix. In DeepSeek-V3 and GPT-4o, when the item order is fixed, the prompted preference appears at both levels. Randomizing item order reduces it in the response geometry, while leaving it in the Big Five scores: the Big Five still separate the two personas at $95$--$97\%$, while response geometry falls to chance.

In randomized administration, the response geometry still contains information about the respondent. In DeepSeek-V4-Flash, SPD features are associated with whether the same instance sorts Asian American public figures as ``foreign'' in a text IAT ($p = .047$), while the Big Five do not. In human studies conducted using randomized item order, the response geometry identifies people from test-retest sessions in $81$ -- $85\%$, against chance $0.4\%$, and predicts sex and, weakly, age. 

In addition, SPD features also predict complex human behavior (e.g., cumulative GPA, $R^2 = 0.28$, permutation $p = .003$), where the Big Five do not.
\end{abstract}

\paragraph{Significance Statement.}
This work shows that cultural personas in LLM survey respondents induce a measurable bias that persists in aggregate Big Five scores. By modeling within-respondent response geometry on the SPD manifold and randomizing item order, we reduce this persona bias, and the geometry that remains is associated with the same instance's subsequent behavior, where the Big Five scores are not. In humans, the same geometry identifies individuals, predicts demographics, and predicts GPA where the Big Five scores do not. These findings provide a measurement framework for using LLMs as survey respondents and suggest that the response geometry captures a stable, behaviorally relevant structure beyond the aggregate scores, in people, and preliminarily in LLMs.

\section{Introduction}
\label{sec:intro}
Cross-cultural research has begun to run on synthetic respondents. Prompt a language model to answer as an American, or as a Chinese-American, or as a member of any group of interest, and it will complete a personality inventory, a values questionnaire, or an attitude survey; the resulting profiles differ between personas in ways that track documented cultural differences, and they are increasingly used in place of human samples on that basis \citep{argyle2023out, dey2025cultural}.

Left unprompted, LLMs answer personality questionnaires like Western, English-speaking populations, a pattern traced to training text and to human feedback in alignment (RLHF) \citep{atari2023which, tao2024cultural}; alignment tuning shifts personality scores directly \citep{lee2025trait}. Prompting a non-Western persona has been proposed as the remedy \citep{tao2024cultural}. But the remedy repeats the mechanism it targets: a cultural preference supplied from outside the instance---first by training, then by the prompt. Neither the default Western lean nor the prompted non-Western one is a trait the questionnaire measures. Both distort the measurement. The bias, then, is relative to the measurement goal, not to an ideal respondent. Scores that carry such a preference can lead to the conclusion that a model's behavior cannot be predicted from what it says about itself. Personality self-reports of language models do not reliably predict their behavior, and persona prompting shifts the self-reports but has a negligible or inconsistent effect on behavior \citep{han2025illusion}. Whether this dissociation reflects the models or the conditions under which coherence is measured remains open \citep{kocielnik2026rethinking}.


At the level of a population, the difference a persona installs is the point; at the level of an instance whose response is read as a measurement, the same difference is a bias that the prompt, not the instance, supplies. Bias in a respondent is rarely something the researcher can change; in cross-cultural research with people, it is handled at the level of the instrument and its scoring \citep{he2012bias}. We take the same stance for models: the model's bias is given, and the scoring is what can be chosen.

A questionnaire response set has two layers: the item responses, which aggregate into dimension scores, and the covariance among items within one response set, which dimensional averaging discards. For a single response set we arrange the responses into an Item--Dimension matrix, compute the correlation matrix across its dimension columns, and represent it as a point on the manifold of symmetric positive definite (SPD) matrices --- the \emph{response geometry}.

Item order matters for this representation. Presenting items in random order is standard practice for removing order effects, and it bears directly on a within-respondent correlation matrix: the matrix pairs the $k$-th item of one dimension with the $k$-th item of another, and that pairing depends on the order used to fill the matrix. Responses collected in random orders can be written back into one common order and analyzed under that shared order; we apply this realignment to both respondent types.

The approach draws on five fields: psychometrics for the instrument and its scoring, Riemannian geometry to represent the correlation matrix of one respondent in the SPD manifold \citep{pennec2006riemannian, arsigny2007logarithmic}, a representation established in functional-connectivity research in network neuroscience \citep{barachant2013classification}, machine learning for unsupervised classification and cross-validated prediction from these representations \citep{hastie2009elements}, social psychology for an implicit-bias criterion \citep{devos2005american}, and the study of language models for the respondents themselves.

We ask two questions. The first is whether the persona bias can be reduced: in DeepSeek-V3 and GPT-4o, we install a cultural persona bias by prompting an American or a Chinese-American persona and test how far randomized administration scored by response geometry reduces it. The second is whether what remains still measures the respondent. We test this in people, where no persona is prompted, by asking whether the response geometry identifies individuals across sessions and predicts sex, age, and GPA; and in a language model (DeepSeek-V4-Flash), by asking whether it is associated with what the same instance does next in an implicit-association task. In both parts the response geometry is compared with the Big Five scores, and both are accompanied by a simulation that keeps every item mean but removes the correlations between items.

\section{Related Work}
\label{sec:related}

\paragraph{Cultural bias and persona prompting in language models.}  Language models are increasingly used as survey respondents and as stand-ins for human samples \citep{argyle2023out, dey2025cultural, lutz2025prompt}. Their responses to psychological measures are more similar to those of Western, educated, industrialized, rich, and democratic populations \citep{atari2023which}, and their responses to cross-cultural value surveys lean toward English-speaking and Protestant European countries \citep{tao2024cultural}. This lean is attributed to the training text that over-represents these populations \citep{atari2023which, tao2024cultural} and to human feedback on alignment \citep{tao2024cultural}; alignment tuning directly shifts personality scores \citep{lee2025trait}. It is a learned cultural preference, not traits that the questionnaire measures, and a persona prompt provides a second preference of the same kind rather than removing the first. Prompting a cultural persona moves the responses of a model closer to those of the prompted country, though not for every country \citep{tao2024cultural}, and the question of whether such survey responses carry information about a model's behavior has been left open \citep{kocielnik2026rethinking}. Survey responses are also sensitive to the order and labeling of options, to the point that corrected responses drift toward uniform \citep{dominguez2024questioning}.

\paragraph{Personality measurement of language models.} For language models, aggregate scores have been found to separate prompted personas and track the norms of human populations \citep{safdari2023personality, jiang2023evaluating}. However, these scores do not reliably measure an LLM. The human factor structure does not replicate in the responses of the pooled models \citep{peereboom2025cognitive, suhr2025challenging}; the personality scores of the model are shaped by training and alignment \citep{lee2025trait}, and they shift when an equivalent prompt or a different order of options is used \citep{gupta2024self}. 

In addition, all of these findings pertain to the aggregate
score: factor loadings are computed from covariances \emph{between}
subjects and are undefined for a single subject, while the score itself
moves with training and with the instrument. None examines the covariance
\emph{within} one subject's response set. The response geometry used here
is defined precisely at that level (Section~\ref{sec:method}).

\paragraph{Bias in psychological measurement.} Cross-cultural psychology treats bias as a property of how an instrument functions between groups and addresses it through instrument design and scoring \citep{he2012bias}. Personnel psychology
faces a related problem: instruments whose scores differ between groups produce an adverse impact, and alternative predictors and scoring choices are sought that reduce subgroup differences while maintaining validity \citep{ployhart2008diversity}.

\paragraph{Item order.} Question-order effects are long documented in human surveys \citep{schuman1996questions}; in language models, order sensitivity has been studied mainly in few-shot prompting \citep{zhao2021calibrate}. Here item order is manipulated to test which layer of the response a persona difference depends on.

\paragraph{Response geometry.} This representation borrows from two traditions. First, from functional connectivity in network neuroscience, it takes the idea that one recording has two complementary descriptions: what each region does on average, and how regions co-vary over time \citep{friston2011functional, sporns2013structure}. Such correlation matrices are compared in the tangent space of the SPD manifold \citep{barachant2013classification,pennec2006riemannian, arsigny2007logarithmic}. In a questionnaire, dimension means play the role of average activations, and correlations among dimension columns play the role of connectivity.

Second, from attention in sequence models, it takes the idea of treating each element of a sequence as a token and describing the sequence by the relations among its tokens \citep{vaswani2017attention}. We treat one questionnaire response as a short sequence in which each item response is a token. In IPIP-50, for example, the scoring key assigns each of the 50 tokens to one of five dimensions (ten per dimension); within each dimension, the tokens are paired by their order, and the correlations between dimensions form a $5\times5$ SPD matrix. Unlike attention in a transformer, these relations are not learned: they are fixed by the scoring key and the item order, and they relate dimensions rather than individual tokens. The sequence enters only through the order of its tokens; no reaction times or internals of the model are used.

\paragraph{Implicit measures and model behavior.} Unlike item scores or dimension scores (i.e., the Big Five scores), the covariance among a respondent's answers is not a quantity the respondent sets item by item; in that sense, the response geometry (i.e., the SPD matrix) resembles implicit measures, which are by-products of responding rather than reports \citep{greenwald1998measuring}. In humans, explicit self-report and implicit measures correlate only weakly \citep{hofmann2005meta, oswald2013predicting}. In language models, self-reports and behavior cohere only for properties anchored outside the immediate prompt, and persona prompting moves what a model says about itself more than what it does \citep{kocielnik2026rethinking, han2025illusion}. The behavioral criterion used here is the American \, $=$ \, White paradigm \citep{devos2005american}.

\section{Method}
\label{sec:method}

\paragraph{Response geometry.} Each IPIP-50 item belongs to one of five dimensions (10 items each). After reverse keying, a respondent's 50 responses are arranged into a $10\times5$ matrix $\mathbf{X}$ whose columns are the dimensions and whose rows pair the $k$-th item of each dimension. We compute the correlation matrix within the respondent

\begin{equation}
\label{eq:corr}
\mathbf{C} = \mathrm{corr}\big(\text{columns of } \mathbf{X}\big) \in \mathbb{R}^{5\times5},
\end{equation}

\noindent map it to the tangent space by the matrix logarithm, and take the 10 upper off-diagonal entries of $\log \mathbf{C}$ as the \textbf{SPD features} (\emph{response geometry}).

\paragraph{Degenerate matrices.} When all items of a dimension receive the same answer, or the correlation matrix $\mathbf{C}$ in Eq.~\eqref{eq:corr} is singular, $\mathbf{C}$ is not strictly positive definite and its logarithm is undefined. This affects $14.0\%$ (DeepSeek-V3) and $11.2\%$ (GPT-4o) of the random-order matrices in the persona experiment and $3.0\%$ in IAT collections 1--3. These matrices are regularized before the logarithm is taken, so that no response is discarded.\footnote{Persona experiment: Gaussian noise is added to a zero-variance column ($\sigma = 0.01$) and to the whole matrix ($\sigma = 10^{-6}$), and $10^{-6}\mathbf{I}$ is added if an eigenvalue remains non-positive. IAT experiment: Gaussian noise ($\sigma = 0.1$) is added to every entry, and the SPD features are averaged over 20 draws.}

\paragraph{Features.} Alongside the SPD features, we also compute the five \textbf{Big Five} dimension means and, as a reference, the reverse-keyed item responses (\textbf{raw items}).

\paragraph{Item order.} The correlation matrix depends on how the items are paired. The row $k$ of $\mathbf{X}$ pairs the $k$-th item of each dimension, and the correlation between two dimensions is computed over these ten pairs. Reordering the items within a dimension changes which items are paired and, therefore, changes the matrix; the Big Five scores, which are column means, do not change. If, for example, the first Extraversion item and the first Agreeableness item share row~1, shuffling the Extraversion column pairs that Agreeableness item with a different Extraversion item.

Under \textbf{fixed order (FO)} every respondent receives the items in the standard IPIP-50 sequence, and the matrix is filled in that sequence. Under \textbf{random order (RO)}, which we call randomized administration, each respondent receives its own random permutation of the items.

The \textbf{RO} responses can be filled into the matrix in three ways: in the order in which the respondent encountered the items (\emph{RO-order}; each respondent has its own item order); realigned to the standard sequence (\emph{RO-canonical}; one order shared by all respondents); or realigned to a randomly drawn within-dimension order shared by all respondents, repeated for 200 draws (\emph{RO-shared}). The responses are reverse keyed before the matrix is filled.

\section{Experimental Setup}
\label{sec:setup}

\subsection{LLM Experiments}
\label{sec:setup_llm}
The LLM experiments install a cultural persona bias
in two models and test whether randomized administration scored by response geometry reduces it (persona experiment), and whether the questionnaire response is associated with the model's behavior (IAT experiment).

\paragraph{Persona.} In the persona experiment, GPT-4o and DeepSeek-V3 answered IPIP-50 \citep{goldberg1999broad} while playing an American or a Chinese-American person. Each response is a separate API call (temperature $0.7$), that is, one independent respondent. The items were given in standard order (fixed order) or in a new random order for each call (random order), with 92--100 responses per model, persona, and order (prompts in SI Appendix, Section~S3).

We ask whether an unsupervised method, which never sees persona labels, recovers the two personas: PCA (five components) followed by $k$-means with two clusters, with clusters matched to personas in the training folds of 10 repetitions of 10-fold cross-validation. We report the percentage of correctly assigned responses (chance $50\%$) and compare the personas on each SPD coordinate (Cohen's $d$). A simulation that keeps every item mean but removes the correlations between items tests whether the geometric separation can arise from the item means alone (SI Appendix, Section~S1).

\paragraph{Behavior (IAT).} The Implicit Association Test (IAT) measures associations that people do not report, from how they classify stimuli into categories \citep{greenwald1998measuring}. With it, \citet{devos2005american} showed that Americans associate ``American'' more with White than with Asian Americans, even for well-known Asian Americans who are United States citizens. In language models, this implicit measure is the one least affected by a self-report given earlier in the same conversation \citep{kocielnik2026rethinking}. We use it as a behavior that follows the questionnaire and ask whether the questionnaire is associated with it. The prompts and names are in the SI Appendix, Section~S4.

In the IAT experiment, DeepSeek-V4-Flash played the Chinese-American person in 300 separate conversations.\footnote{The IAT experiment used the same \texttt{deepseek-chat} endpoint as the persona experiment; from April 24, 2026, the provider routed this endpoint to DeepSeek-V4-Flash \citep{deepseek2026v4}, so the persona experiment (March 2026) ran on DeepSeek-V3 and the IAT experiment (September 2026) on DeepSeek-V4-Flash.} In each, it first answered the IPIP-50 in random order (realigned to the standard order for the SPD features), then labeled eight well-known United States citizens --- four White and four Asian American --- as ``American'' or ``Foreign''. The outcome of each sorting trial is coded 1 if it labeled all four White American names ``American'' and all four Asian American names ``Foreign'', and 0 otherwise.

The persona prompt, the IPIP-50, and the IAT were given within one conversation, because an LLM instance exists only within a single conversation. Linking the two at the instance level therefore requires measuring the questionnaire and the behavior in the same conversation. Conversations were collected in five collections ($N = 50$, $50$, $100$, $50$, and $50$). In collections~4 and~5, the sorting trial was repeated ten times in each conversation, each repeat continuing from the same IPIP-50 exchange, which gives 1{,}200 trials in total (205 stereotyped).\footnote{In collection~1 the IPIP-50 response scale ran in the opposite direction (A = Very Inaccurate). Collection~2 also included 50 instances with the American persona; none of them produced the stereotyped sorting ($0/50$, against $15/50$ under the Chinese-American persona).}

All collections were analyzed together, with features standardized within collection and the collection entered as a fixed effect. Because collections~4 and~5 contain repeated trials, the outcome was modeled by mixed-effects logistic regression \citep{bates2015fitting} with a random intercept for each instance, and each feature set was tested by a likelihood ratio against the model with collection only. The feature sets are the complete SPD features (10d), the five Big Five scores (5d), and the C--O coordinate of the SPD features (1d, selected in the first collection, where it separated instances by sorting behavior, and tested one-sided in that direction). The other nine SPD coordinates were also tested individually (two-sided, uncorrected; SI Appendix, Section~S4). Results by collection are in SI Appendix, Section~S4.

\subsection{Human Comparison}
\label{sec:setup_human}
The human comparison tests whether response geometry is a meaningful measurement when no persona is prompted: whether it is stable across sessions and predicts behavior and demographic variables, compared with the Big Five scores. We used two data sets.

\paragraph{NEO-PI-R (public data).} A public data set of 263 adults who completed the NEO-PI-R twice, one week apart, each with the items in random order \citep{henry2026neo, henry2026neodata}; it also records age and sex. We ask whether a person can be recognized from their responses: each person's first-session correlation matrix is compared with all 263 second-session matrices, and recognition succeeds when the most similar one is their own (chance $1/263$), with the items realigned to the standard order or to 200 random shared orders. With eight items per facet, each $30\times30$ matrix has rank at most 7 and is singular, so the matrices are compared entry by entry. To predict age and sex, each first-session matrix is first made positive definite by Ledoit--Wolf shrinkage toward the identity \citep{ledoit2004well} and then mapped to the tangent space, and predictions are evaluated by cross-validation with permutation tests.\footnote{The shrinkage coefficient is estimated per respondent (median $0.785$, range $0.373$--$0.876$). Age is predicted by elastic-net \citep{zou2005regularization} and ridge regression, sex by $L_1$- and $L_2$-penalized logistic regression. Predictors are standardized and penalties tuned by 5-fold cross-validation within each training fold; performance is the out-of-fold $R^2$ or AUC from 10-fold cross-validation, averaged over three fold assignments. The lasso is not used for age because it retained no predictor in 7 of 30 folds.}

\paragraph{GPA (private data).} Eighty-nine undergraduates completed an
IPIP-100 in a split-half design across two sessions within one semester,
several days to weeks apart. The inventory has five dimensions. Each session
contained a different half of the items (50 items, 10 per dimension), with
its own scoring key. Within each session the items were in a fixed order; the two sessions differed in both the items and their order, and the two scoring keys were independent. GPA is predicted from Session~1 only, which was completed under standard instructions, because Session~2 was administered under a laboratory manipulation (a simulated selection condition) for a different purpose \citep{yuan20xxthesis}.

The cumulative GPA was obtained from university records. We
predict GPA from the Session-1 SPD features and, for comparison, from the Big
Five scores computed from the same Session-1 responses. Data were collected for a master's thesis \citep{yuan20xxthesis}; the response-geometry analyses reported here are new.

\section{Results}
\label{sec:results}

\subsection{LLM Experiments}

\begin{figure}[htbp]
\centering
\includegraphics[width=.75\linewidth]{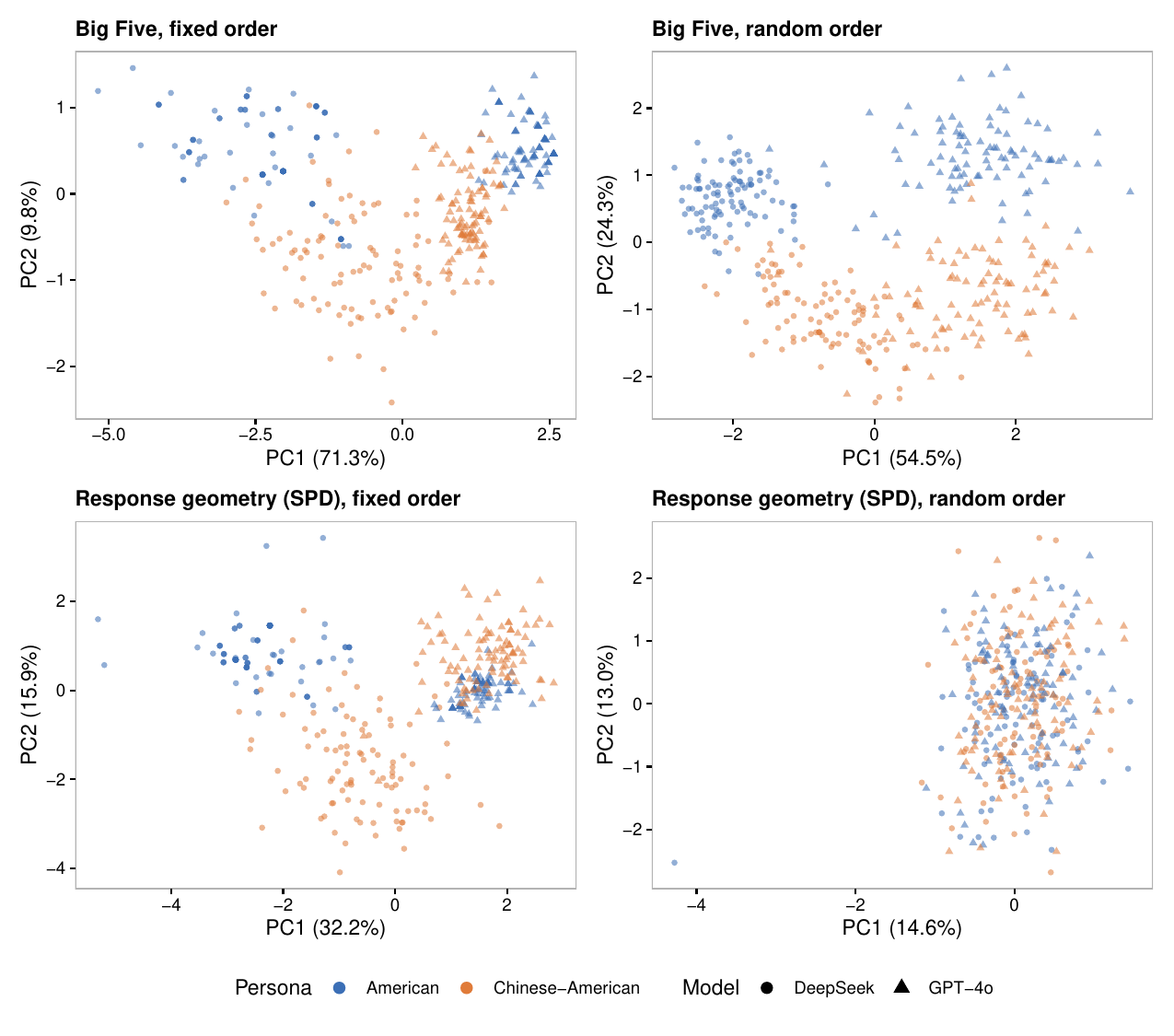}
\caption{Persona separation under fixed and random item order. Each panel shows the first two principal components of a PCA fitted to both models (circles, DeepSeek-V3; triangles, GPT-4o); random order is realigned by key. Persona classification is evaluated within model in Table~\ref{tab:llm}.}
\label{fig:pca}
\end{figure}

\paragraph{Persona.} In fixed order, both the Big Five scores and the response geometry carry the prompted cultural bias (Table~\ref{tab:llm}).
The Big Five scores separate the two personas with accuracies of $79.45\%$ (DeepSeek-V3) and $93.27\%$ (GPT-4o); the response geometry separates them at $94.80\%$ and $94.99\%$.\footnote{The geometric separation spans several SPD coordinates: $7$ of $10$ (DeepSeek-V3) and $6$ of $10$ (GPT-4o) differ between personas at $p<.01$ under fixed order, against $0$--$2$ of $10$ under random order. It is not explained by item means alone: a simulation that keeps GPT-4o's item means and SDs but draws items independently reproduces the Big Five separation ($97.6$--$98.3\%$) but not the geometric separation ($62.6$--$64.0\%$, against $95.0\%$ measured).}
Figure~\ref{fig:pca} shows the same pattern: the personas separate in Big Five scores under both orders and in SPD features under fixed order, but not in SPD features under random order.

\begin{table}[htbp]
\small
\centering
\caption{Persona separation (unsupervised clustering accuracy, \%, mean $\pm$
SD over 10 repetitions of 10-fold CV).}
\label{tab:llm}
\setlength{\tabcolsep}{3pt}
\begin{tabular}{llccc}
\toprule
& & \textbf{FO} & \textbf{RO-canonical} & \textbf{RO-order} \\
\midrule
DeepSeek-V3 & Raw items & 100.00 $\pm$ 0.00 & 98.55 $\pm$ 0.37 & 98.55 $\pm$ 0.37 \\
            & Big Five  & 79.45 $\pm$ 1.21 & 96.60 $\pm$ 0.21 & 96.60 $\pm$ 0.21 \\
            & SPD       & \textbf{94.80} $\pm$ 0.26 & \textbf{50.00} $\pm$ 0.00 & 49.90 $\pm$ 1.98 \\
\addlinespace[2pt]
GPT-4o      & Raw items & 100.00 $\pm$ 0.00 & 98.98 $\pm$ 0.65 & 98.98 $\pm$ 0.65 \\
            & Big Five  & 93.27 $\pm$ 0.04 & 95.10 $\pm$ 0.24 & 95.10 $\pm$ 0.24 \\
            & SPD       & \textbf{94.99} $\pm$ 0.47 & \textbf{48.51} $\pm$ 0.70 & 48.57 $\pm$ 3.98 \\
\bottomrule
\end{tabular}
\vspace{2pt}\\
\footnotesize
\textbf{Columns.} \textit{FO} = fixed order. \textit{RO-canonical} = random order, realigned to the standard item sequence so that all instances share one order. \textit{RO-order} = random order, with each instance's matrix filled in its own encounter order. Here \textit{order} means item presentation order. \textbf{RO-shared.} Realigned to 200 random shared orders, SPD separation is $50.13 \pm 3.41$ (DeepSeek-V3) and $49.54 \pm 4.27$ (GPT-4o).
\end{table}

\paragraph{Persona under random order.} Under random order, the Big Five
scores still carry the prompted culture: they separate the two personas at $96.60\%$ (DeepSeek-V3) and $95.10\%$ (GPT-4o; Table~\ref{tab:llm}). Therefore, at the aggregate score level, randomizing the order of the items does not remove persona bias.

The keyed response geometry behaves differently. When each instance's matrix is filled in its own encounter order (RO-order), persona separation falls to $49.90\%$ (DeepSeek-V3) and $48.57\%$ (GPT-4o), which is chance. Realigning the responses to a shared item order (RO-canonical) does not recover the separation: $50.00\%$ and $48.51\%$, again chance.

This contrast is the main result of the persona experiment. Under random order, the prompted culture remains detectable in the Big Five scores but not in the keyed geometry. Raw-item and Big Five features do not depend on the order of the items, so their two RO columns in Table~\ref{tab:llm} are identical; only the SPD geometry depends on the order in which the items are encountered.

Chance-level separation does not mean that the keyed matrices are empty. With ten rows per dimension, two unrelated columns give a mean absolute correlation of about $0.27$, and this is what the matrices show when each instance is filled in its own encounter order ($0.27$ in both models). Under a shared order (RO-canonical), the correlations are larger: $0.29$ (DeepSeek-V3) and $0.41$ (GPT-4o; paired Wilcoxon test against RO-order, $p = .004$ and $p < .001$; SI Appendix, Table~S9). Under randomized administration, the keyed geometry therefore retains structure, but this structure no longer separates the personas.

\paragraph{Behavior (IAT).} Across the five collections (300 instances, 1{,}200 trials, 205 stereotyped), the ten SPD features together are associated with the stereotyped sorting ($p = .047$), the C--O coordinate is associated in the expected direction at the margin (one-sided $p = .051$), the N--O coordinate shows the strongest single association ($p = .011$, uncorrected), and the Big Five scores are not ($p = .20$; Table~\ref{tab:iat}). The results for each collection, including the consistency of classification in repeated trials of collections~4 and~5, are in the SI appendix, Section~S4.

\begin{table}[H]
\centering
\caption{Stereotyped sorting in the IAT task and the IPIP-50 response (300 instances, 1{,}200 trials, 205 stereotyped).}
\label{tab:iat}
\small
\setlength{\tabcolsep}{4pt}
\begin{tabular}{lcc}
\toprule
\textbf{Features} & \textbf{Test} & \textbf{$p$} \\
\midrule
SPD, C--O coordinate & $\beta=-0.17$ & $.051^{\dagger}$ \\
SPD, N--O coordinate & $\beta=-0.26$ & $.011^{*}$ \\
SPD, 10 coordinates & $\chi^2(10)=18.5$ & $.047^{*}$ \\
Big Five, 5 dimensions & $\chi^2(5)=7.3$ & $.20$ \\
\bottomrule
\end{tabular}
\vspace{2pt}

\footnotesize
Mixed-effects logistic regression with collection as a fixed effect and a
random intercept per instance. Feature sets
tested by likelihood ratio against the model with collection only; C--O
(per SD) one-sided in the direction observed in the first collection; N--O
(per SD) two-sided, not corrected for the ten coordinates.
$^{*}p<.05$, $^{\dagger}p<.10$.
\end{table}

\subsection{Human Comparison}

\paragraph{NEO-PI-R (public data).} We next asked whether the response geometry
identifies individuals in human data. In the NEO-PI-R sample, 263 adults
completed the questionnaire twice, one week apart, each time in an individually
random order of items. We compared each person's first-session correlation matrix
with all 263 second-session matrices and counted a success when the most similar
matrix was their own (chance $1/263 = 0.38\%$).

After realignment with the standard item order, the geometry identifies individuals well above chance: under reverse-keyed scoring, the nearest-neighbor accuracy is $81.0\%$ with the diagonal included and $79.7\%$ without it; under raw scoring, it is $84.8\%$ and $82.3\%$, respectively (Table~\ref{tab:neo}).

Note that such identification does not depend on the order chosen for realignment: across 200 random shared orders, it is $79.9 \pm 1.3\%$ (reverse keyed) and $86.7 \pm 0.9\%$ (raw). Again, it cannot be reproduced from facet levels alone: in a simulation from the sample's own item statistics, in which each person has stable facet levels but no correlation structure of their own, facet means identify $99.6\%$ of persons but the correlation matrices only $0.25\%$, against $80.8\%$ in the real responses.

\begin{table}[H]
\centering
\caption{NEO-PI-R test--retest data \citep{henry2026neo} ($N=263$, one week,
$30\times30$ facet matrices). Fingerprint = nearest-neighbour identification
accuracy; chance $=0.38\%$.}
\label{tab:neo}
\small
\setlength{\tabcolsep}{3.5pt}
\begin{tabular}{llcccc}
\toprule
\textbf{Scoring} & \textbf{Tri.} & \textbf{Within} & \textbf{Between} & \textbf{$w{>}b$} & \textbf{Fingerprint} \\
\midrule
Reverse-keyed & +diag & 0.542 & 0.336 & 98.1\% & 81.0\% \\
Reverse-keyed & off & 0.339 & 0.043 & 98.1\% & 79.7\% \\
Raw & +diag & 0.695 & 0.329 & 99.6\% & 84.8\% \\
Raw & off & 0.581 & 0.093 & 98.9\% & 82.3\% \\
\bottomrule
\end{tabular}
\vspace{2pt}

\footnotesize
Within/Between = mean element-wise Pearson $r$ for matched/mismatched session pairs; $w{>}b$ = fraction of participants whose within exceeds their mean between. All four cells: permutation $p = .005$, the floor with 200 label permutations.
\end{table}

The same geometry also predicts demographic variables. In the NEO-PI-R sample, response geometry predicts sex with cross-validated AUC $0.73$--$0.75$ (permutation $p<.005$; Table~\ref{tab:neo_agesex}). It is associated with age in the whole sample (permutation $p \le .010$), but predicts age only weakly out of sample (cross-validated $R^2 = 0.03$--$0.05$).\footnote{With the 435 untransformed correlations as features, the pattern is the same: sex AUC $0.70$ ($L_1$) and $0.75$ ($L_2$), age $R^2 = 0.08$ (ridge).}

\begin{table}[H]
\centering
\caption{Prediction of age and sex from Session-1 NEO-PI-R responses ($N=263$).}
\label{tab:neo_agesex}
\small
\setlength{\tabcolsep}{3.5pt}
\begin{tabular}{lcccc}
\toprule
 & \multicolumn{2}{c}{\textbf{Age}} & \multicolumn{2}{c}{\textbf{Sex}} \\
\cmidrule(lr){2-3}\cmidrule(lr){4-5}
\textbf{Predictor set} & \textbf{CV $R^2$ (EN / Ridge)} & \textbf{$p$} & \textbf{CV AUC ($L_1$ / $L_2$)} & \textbf{$p$} \\
\midrule
Facet means (30) & 0.191 / 0.197 & $<.005$ / $<.005$ & 0.744 / 0.730 & $<.005$ / $<.005$ \\
Response geometry (435) & 0.029 / 0.048 & $.010$ / $<.005$$^{\ddagger}$ & 0.727 / 0.745 & $<.005$ / $<.005$ \\
\bottomrule
\end{tabular}
\vspace{2pt}

\footnotesize
EN: elastic-net regression (a mix of lasso and ridge penalties). Ridge: ridge
regression. $L_1$: lasso-penalized logistic regression. $L_2$: ridge-penalized
logistic regression. 10-fold cross-validation, three fold assignments. $p$: 200
permutations ($.005$ is the floor). $^{\ddagger}$Test of the in-sample fit ($R^2 = 0.355$ / $0.488$).
\end{table}

\paragraph{GPA (private data).} In a sample of 89 college students, the SPD features of the Session-1 matrix predict cumulative GPA ($R^2 = 0.281$, permutation $p = .003$; Table~\ref{tab:gpa}), while the Big Five scores of the same responses do not ($R^2 = 0.018$, $p = .91$). This mirrors the LLM result: the geometry carries information that the aggregate scores do not.

\begin{table}[H]
\centering
\caption{Prediction of cumulative GPA from Session-1 responses ($N{=}89$).}
\label{tab:gpa}
\small
\setlength{\tabcolsep}{4pt}
\begin{tabular}{lccc}
\toprule
\textbf{Predictor set} & \textbf{$R^2$} & \textbf{Adjusted $R^2$} & \textbf{10-fold CV $R^2$} \\
\midrule
SPD tangent space (10D) & 0.281$^{**}$ & 0.188 & 0.183 \\
Big Five (5D) & 0.018 & $-$0.041 & 0.110 \\
\bottomrule
\end{tabular}
\vspace{2pt}

\footnotesize
Significance of $R^2$ by permutation test (GPA shuffled across participants):
$^{**}p<.01$ (SPD tangent $p=.003$; Big Five $p=.91$). Untransformed
correlations: $R^2 = 0.213$, $p = .034$. Under repeated 50/50 holdout the ordering is preserved but both
estimates are small ($0.085 \pm 0.062$ vs.\ $0.029 \pm 0.035$).
\end{table}

\section{Discussion}

\paragraph{The structure of the bias.} Across the LLM experiments, the cultural bias induced by persona prompting appears at two levels. In fixed order, the bias is present in both the Big Five scores and the keyed response geometry, where it cannot be reproduced from the item means. Under random order, the bias remains at only one level: the Big Five scores still carry it, the keyed geometry does not, and realignment to a shared order does not recover it.

\paragraph{Reducing bias.} Randomized administration read through the keyed
response geometry reduces persona bias. In the LLM persona experiment, the raw items of both models separate the two personas almost perfectly ($98.6$--$99.0\%$), and the Big Five scores retain most of this separation ($95.1$--$96.6\%$), while the keyed geometry is at chance ($48.5$--$50.1\%$).
In the LLM IAT experiment, the reduction was only partial: in collection~2, where both personas were prompted, the keyed geometry still separated them at $67\%$, against $100\%$ for the Big Five scores (SI Appendix, Section~S4).
Across both experiments, the keyed geometry carried less of the persona bias than the item or Big Five scores.

\paragraph{Cultural research using language models.} When models serve as
subjects of cultural inquiry, differences in Big Five scores induced by persona
prompts can easily be read as cultural differences. LLMs carry biases of their
own, and persona prompts can add group-level differences on top of them; whether
such differences appear depends on the prompt and on how the responses are
scored. Self-report scores, for humans and for LLMs alike, mix stable traits
with response biases, and the questionnaire alone cannot separate the two.
Randomizing item order and scoring the keyed response geometry reduces the
variance installed by the persona.

This matters because the variance removed does not describe the respondent. In
human assessment, score differences between groups have an adverse impact, and
personnel psychology has long sought scoring methods that reduce subgroup
differences while preserving validity \citep{ployhart2008diversity}. In people,
subgroup differences may be partly real; here, the persona differences are
installed by the prompt, so reducing them need not cost information about the
individual instance. The keyed response geometry under randomized administration is
one such scoring method.

\paragraph{Instance-level link.}
An IAT effect in an LLM is not itself rare \citep{bai2025explicitly}. What is
rare is instance-level evidence: evidence that one instance's responses to
ordinary personality items carry information about that same instance's
subsequent behavior, which self-reports on such inventories have not been found
to provide \citep{han2025illusion, kocielnik2026rethinking}.

Our IAT experiment provides preliminary evidence for such a link. Across all
collections, the ten SPD features together are associated with the stereotyped
sorting, and the C--O coordinate is associated in the expected direction at the
margin, while the Big Five scores are not. The
contribution is not the existence of an IAT effect, but the instance-level
link from questionnaire response to behavior.

We claim only an association. Out-of-sample prediction would require a different
experimental design, because the behavior of a language model fluctuates
considerably even across repeated trials of the same instance (SI Appendix,
Section~S4). The response geometry may have the potential for out-of-sample
prediction, but whether it can actually predict depends on the consistency of the
model's behavior and on how much valid variance that behavior contains.

\paragraph{Within-instance design.}
In LLMs, explicit responses do not reveal implicit associations \citep{bai2025explicitly}. We therefore measured the questionnaire and the behavior within the same instance. \citet{bai2025explicitly} also paired two tasks within one prompt, but their association task named the social groups and attributes immediately before the decision task, so the measure itself could cue the behavior it is paired with. In our design, all instances received the same Chinese-American persona prompt and answered ordinary personality items before the sorting task; the questionnaire contains no group or stereotype content, and the predictor is a by-product of responding rather than a response about the groups. Under these conditions, the SPD features together were associated with stereotyped sorting, whereas the Big Five scores were not. A shared context alone would be expected to affect both feature sets, so it does not explain why only the SPD features were associated with it. The link is better attributed to how the instance arranges its responses across items.

\paragraph{From bias to measurement.} Reducing persona bias does not leave the measurement empty. In people, the same keyed geometry identifies individuals after randomized administration, predicts sex, and predicts GPA where the Big Five scores do not. In the LLM IAT experiment, it is weakly associated with what the same instance does next. Measures whose scores predict what a respondent does are needed for people and language models alike, and developing them remains an open task; response geometry read under randomized administration is one step in that direction.

\section{Limitations}

\paragraph{From procedure to measurement.}
This paper offers a scoring procedure --- randomized administration scored by the keyed response geometry --- that reduces persona bias in the two models tested. Its value as a measurement currently rests on human data and preliminary behavioral evidence from models. More human samples, models, cultural personas, and behavioral criteria are needed to fully establish the procedure as a measurement.

\paragraph{Scope.}
Three inventories are used throughout the study: IPIP-50 in the LLM
experiments and NEO-PI-R and an IPIP-100 split-half administration in
the human comparison. The persona experiment uses two models (DeepSeek-V3 and GPT-4o), two
personas, and one inventory (IPIP-50, whose $10\times5$ structure the method
requires): a minimal contrast. The IAT experiment uses a third model, DeepSeek-V4-Flash, in which the response geometry also partially reduced persona bias (collection~2; SI Appendix, Section~S4). The two persona-experiment models received different
prompts (a numeric scale for DeepSeek-V3, a letter scale for GPT-4o, with
different persona wording), and all models were accessed through unpinned
endpoints. The NEO-PI-R analysis is performed at the facet level, and the analysis for LLM experiments is performed at the dimension level.

\paragraph{Limited validity of the IAT criterion.} Bias is itself a behavior:
it appears in the questionnaire responses as persona separation and in the IAT
as stereotyped sorting, and the two need not agree. The IAT also has known
limitations as a behavioral criterion. In people, meta-analyses find only a weak
prediction of behavior from IAT scores ($r \approx .15$--$.27$;
\citealp{greenwald2009understanding, oswald2013predicting}), and reviews
conclude that implicit measures add little beyond self-report
\citep{meissner2019predicting}. In LLMs, IAT effects are common
\citep{bai2025explicitly}, but implicit associations do not predict behavior
between models \citep{andric2025walk}. In addition, within one persona, instances
differ only through sampling, which leaves little variance in the criterion to
predict. These limitations make a weak association unsurprising, and part of its
weakness may reflect the criterion rather than the questionnaire.

\section{Conclusion}
Research on culture in language models has mostly asked what a model says: how high its scores are and how they differ between prompted groups. Our results suggest that this is the part of a response that a prompt can control most easily. A persona can tell a model what to answer, but it has less control over how the answers are arranged across items. Presenting the items in a different random order to each respondent exposes this difference, and response geometry captures it. Evaluating cultural bias in LLMs should therefore look beyond the content of responses to their structure.

The same shift has a long history in psychology. Classical scoring treats within-person covariance as noise around a trait level. Language models make the level the least trustworthy part of a response, because it can be set by a single prompt. In this setting, the structure becomes the informative layer: in people it identifies individuals and predicts GPA, and in a language model it begins to track what the same instance does. Response geometry thus offers a candidate common language for measuring people and models, one that does not depend on what either of them chooses to report.

\section*{Data Availability}
The NEO-PI-R test--retest data are publicly available from the Open Science Framework (\url{https://osf.io/wz3du/}; DOI: \href{https://doi.org/10.17605/OSF.IO/WZ3DU}{10.17605/OSF.IO/WZ3DU}) \citep{henry2026neodata}. The Supplemental Information (SI) Appendix contains the full prompts and IPIP-50 items for both LLM experiments, the data collection details (model endpoints, dates and software), the regularization procedures, the simulation procedures and results, and the extended analyses for the persona, IAT, and NEO-PI-R data. The LLM responses will be released upon acceptance.
GPA data include academic records of students and cannot be made publicly available under the terms of Auburn University IRB approval (protocol \#13-282 MR 1308). However, interested readers may apply to access de-identified data, subject to Auburn University's permission.

\section*{Ethics.} The GPA study was approved by the Auburn University Institutional Review Board, protocol \#13-282 MR 1308, and informed consent was obtained from all participants. The NEO-PI-R data are publicly available (\url{https://osf.io/wz3du/}; DOI: \href{https://doi.org/10.17605/OSF.IO/WZ3DU}{10.17605/OSF.IO/WZ3DU}) \citep{henry2026neodata}. They were collected with the approval of the Research Ethics Committee of the University of Edinburgh School of Philosophy, Psychology, and Language Sciences \citep{henry2026neo}.

\section*{Competing Interest Statement.} The author declares no competing interests.

\section*{Funding.} The author declares no funding.

\section*{Author Contributions.} Y.Y.: conceptualization, methodology, data collection, formal analysis, writing -- original draft, writing -- review \& editing.

\section*{Preprint.} An earlier version of this work was published as a preprint (arXiv:2607.02368).

\FloatBarrier
\bibliographystyle{abbrvnat}
\bibliography{pnas_persona_refs_1}

\end{document}